\documentclass{article}

\usepackage{microtype}
\usepackage{graphicx}
\usepackage{subfigure}
\usepackage{booktabs} 

\usepackage{hyperref}

\usepackage[accepted]{icml2024}

\usepackage{amsmath}
\usepackage{amssymb}
\usepackage{mathtools}
\usepackage{amsthm}

\usepackage[capitalize,noabbrev]{cleveref}

\theoremstyle{plain}
\newtheorem{theorem}{Theorem}[section]

\theoremstyle{definition}
\newtheorem{definition}[theorem]{Definition}

\theoremstyle{remark}

\usepackage[textsize=tiny]{todonotes}

\icmltitlerunning{Event Detection in Time Series: Universal Deep Learning Approach}

\begin{document}

\twocolumn[
\icmltitle{Event Detection in Time Series: Universal Deep Learning Approach}



\icmlsetsymbol{equal}{*}

\begin{icmlauthorlist}
\icmlauthor{Menouar Azib}{}
\icmlauthor{Benjamin Renard}{}
\icmlauthor{Philippe Garnier}{}
\icmlauthor{Vincent Génot}{}
\icmlauthor{Nicolas André}{}
\end{icmlauthorlist}

\icmlkeywords{Machine Learning, ICML}

\vskip 0.3in
]




\begin{abstract}
Event detection in time series is a challenging task due to the prevalence of imbalanced datasets, rare events, and time interval-defined events. Traditional supervised deep learning methods primarily employ binary classification, where each time step is assigned a binary label indicating the presence or absence of an event. However, these methods struggle to handle these specific scenarios effectively. To address these limitations, we propose a novel supervised regression-based deep learning approach that offers several advantages over classification-based methods. Our approach, with a limited number of parameters, can effectively handle various types of events within a unified framework, including rare events and imbalanced datasets. We provide theoretical justifications for its universality and precision and demonstrate its superior performance across diverse domains, particularly for rare events and imbalanced datasets.
\end{abstract}

\section{Introduction}
\label{introduction}
Event detection in time series data is a crucial task in various domains, including finance, healthcare, cybersecurity, and science. This task involves identifying instances of behavioral shifts, often referred to as the change-point detection problem in statistical literature \cite{10.1145/312129.312190}. Such events encompass anomalies, frauds, physical occurrences, and more. In recent years, supervised deep learning methods have emerged as powerful tools for addressing this challenge, often employing a classification framework to assign binary labels to each time step, indicating the presence or absence of an event \cite{9523565, darban2022deep}.

However, these classification-based approaches face inherent limitations. Particularly, they may struggle to handle imbalanced datasets, where the majority class (non-events) significantly outnumbers the minority class (events) \cite{10293442, 10.1145/3292500.3330896}. This imbalance can lead to biased predictions, favoring the majority class and hindering the accurate detection of events. To address this issue, various techniques have been proposed, such as the SMOTE algorithm, which artificially inflates the minority class to improve classification performance. However, these methods have limitations, including potential overfitting and the introduction of artificial data points \cite{Krawczyk2016-tt}.

Moreover, these classification-based methods often fail to consider events defined by time intervals, a common occurrence in real-world scenarios. Events may span multiple time steps, and their identification requires capturing the temporal context of the data rather than simply assigning binary labels to individual time steps. 

Moreover, these methods can be broadly categorized into two approaches: empirical methods primarily focused on practical applications and benchmark performance, lacking a strong theoretical foundation \cite{Ji2021, Rong2022-yi, Reyana2022-id, Mobtahej2022-ny, Gopali2021-ey, darban2022deep, Chalapathy2019-il, Choi2021-mz, Su2019-gk, Li2018-cr, Xu2020-vh, Bashar2020-zg, Han2022-yf, 10.1007/978-3-031-35644-5-17, Du2022-no, 10.1007/978-3-319-24489-1-24, Aminikhanghahi2017-lw, Gupta2022-qd}, and methods with a theoretical foundation grounded in mathematical proofs or justifications for their efficacy. To our knowledge, only one work, \cite{li2023automatic}, falls into the latter category.

In \cite{li2023automatic}, the authors propose a novel change point detection method employing a neural network architecture that includes the CUSUM-based classifier \cite{granjon-hal-00914697} as a specific instance. They demonstrate that their architecture cannot underperform the CUSUM classifier in identifying change points. Additionally, they show that the misclassification error is constrained by two factors: one associated with the inherent misclassification error of the CUSUM-based classifier and the other related to the complexity of the neural network class as measured by its Vapnik-Chervonenkis (VC) dimension \cite{JMLR:v20:17-612}. However, their study has certain limitations. It models the problem as a change-in-mean model \cite{Jewell2022-mv}, which may not be adequately generalized for other types of event detection, including anomalies. Furthermore, it assumes that the data are drawn from a multivariate normal distribution, potentially limiting the applicability of their approach.

In contrast to these limitations, we present a novel supervised deep learning approach for event detection in multivariate time series data that departs from binary classification and leverages regression. This departure offers several advantages in handling imbalanced datasets. Unlike traditional classification, regression inherently accommodates continuous outputs, making it suitable for scenarios where events may not be binary and imbalanced. Additionally, our approach, previously introduced in \cite{azib2023comprehensive}, accepts ground truth events defined as specific time points or intervals, eliminating the need for point-wise labels across the entire dataset. While \cite{azib2023comprehensive} focused on algorithmic implementation and introduced a Python package, our paper delves deeper into the theoretical underpinnings of the method by presenting a mathematical framework.

We demonstrate the universality of our approach in detecting events in time series data, assuming mild continuity assumptions for multivariate time series. By utilizing the universal approximation theorem \cite{Hornik1989-dg}, we establish that our method can detect a broad spectrum of events with arbitrary precision. Notably, our approach surpasses \cite{li2023automatic} in robustness and applicability, owing to its weaker assumptions. This nuanced approach enhances versatility, making it suitable for diverse event detection scenarios. Beyond theoretical considerations, we showcase the practical effectiveness of our approach. Despite having a minimal number of trainable parameters, it outperforms existing deep-learning techniques when applied to real-world imbalanced datasets, such as those in fraud detection and bow shock crossing identification. These empirical validations underscore not only the efficacy but also the broad applicability of our framework across various domains, positioning it as a formidable contender in the field of event detection. The regression-based method, by design, provides a more flexible and nuanced approach to handling imbalanced datasets, contributing to its effectiveness in capturing rare events and continuous variations in event characteristics.

In summary, our proposed framework, rooted in deep learning and regression, offers a robust and versatile solution for event detection in multivariate time series data, particularly in scenarios with imbalanced datasets and non-binary events. It demonstrates superior performance compared to existing methods, both theoretically and empirically. This novel approach holds significant promise for addressing event detection challenges across diverse domains, making it a valuable contribution to the field of time series analysis.

\section{Mathematical Formulation \label{mathematicalFormulation}}
This section introduces the mathematical formulation of the method. The method is based on two key components: a multivariate time series that represents the data, and a set of reference events, also known as ground truth events.
Let $T(t)$ be a time series that maps a real value $t$ to a feature vector in $\mathcal{X} \subset \mathbb{R}^f$, where $f$ is the number of features. The mapping can be represented as follows:
\begin{align*}
T: \mathbb{R} &\rightarrow \mathcal{X} \subset \mathbb{R}^f \\
t &\mapsto T(t)
\end{align*}
The ground truth events are encapsulated within a set $E$. Each event $e$ within this set is an interval defined by a start time $\tau_1$ and an end time $\tau_2$, denoted as:
$$e = [\tau_{1}, \tau_{2}]$$
where both $\tau_{1}$ and $\tau_{2}$ are real numbers.

We assume, without loss of generality, that the time series $T(t)$ is defined for all $t$ in the interval $[\alpha, \beta]$, where $\alpha$ and $\beta$ are real numbers, and that it takes values in $\mathcal{X}$. Additionally, we assume that the reference events in the set $E$ do not overlap, i.e., for all $e_1, e_2 \in E$, the intersection of $e_1$ and $e_2$ is an empty set ($e_1 \cap e_2 = \emptyset$).

\subsection{Overlapping Partitions}
In this subsection, we introduce overlapping partitions concept. We split the interval $[\alpha, \beta]$ into a sequence of equally spaced points $(t_1=\alpha < t_2 < t_3, \ldots < t_N=\beta)$, where $N \in \mathbb{N}$, and the spacing is $s\in \mathbb{R}$.
We consider a family of overlapping partitions $(p_i)_{i \in I}$ where $I = \{i \in \mathbb{N} | 1 \leq i \land i \leq N - w + 1\}$ is an index set. An overlapping partition $p_i$ is set of size $w \in \mathbb{N}, w>1$, denoted by $\{t_i, t_{i+1}, \ldots, t_{i+w-1}\}$, where $i$ is in $I$. The set of all overlapping partitions is $\mathcal{P} = \{p_i | i \in I\}$. The term 'overlapping' means that any two neighboring partitions have at least one point in common.
Next, we define a function $o$ that maps each overlapping partition $p_i \in \mathcal{P}$ to a vector $v_i \in \mathcal{V}$, as follows:
\begin{align*}
o : \mathcal{P} &\rightarrow  \mathcal{V} \\
p_i &\mapsto v_i = o(p_i)
\end{align*}
Here $\mathcal{V}=  \{o(p_i) | i \in I\}$. The function $o$ assigns to the partition $p_i$ a vector $v_i$ of dimension $r = w\cdot f$, which contains the values of the time series $T(t_j)$ for all $j$ such that $i \leq j \land j \leq i+w-1$.
The vector $v_i$ can be written as follows:
\[
v_i = \begin{pmatrix}
T(t_i)[1] \\
\vdots \\
T(t_i)[f] \\
\vdots \\
T(t_{i+w-1})[1] \\
\vdots \\
T(t_{i+w-1})[f] \\
\end{pmatrix}
\]
This representation concatenates the feature values of the time series $T$ over the partition $p_i$, where $T(t_j)[k]$ denotes the $k$-th feature value at time $t_j$.

\subsection{Overlapping Parameter Function}
In this subsection, we introduce the overlapping parameter function, denoted as $op$, which quantifies the temporal distance of each partition $p_i \in \mathcal{P}$ with respect to the nearest events. The function $op$ assigns a value between 0 and 1 to each partition.

To calculate the $op$ value for a given event $e \in E$ and partition $p_i \in \mathcal{P}$, we use the Jaccard similarity coefficient \cite{Jaccard1901-jw}. The $op$ value is computed by taking the duration of the intersection between $p_i$ and $e$, and dividing it by the duration of their union. This is shown in the following formula:
\begin{equation}
\label{eq:1}
op(p_i, e) = \frac{{\text{duration}(p_i \cap e)}}{{\text{duration}(p_i \cup e)}}
\end{equation}

The value of $op(p_i, e)$ will be close to 1 if the event $e$ and the partition $p_i$ largely overlap, and close to 0 if they have little overlap. This provides a measure of the temporal proximity of the partition to the event.

Given that the cardinality of each partition is denoted as $w$, we define the temporal duration $w_s$ as $w_s = (w - 1)\cdot s$. To synchronize the temporal duration of each partition ($w_s$) with the events, we adjust each event $e$ using the following formula:
\begin{align*}
\forall e=[\tau_{1}, \tau_{2}] \in E, \\
t_{mid} = \frac{\tau_{1} + \tau_{2}}{2}, \\
e = \left[\tau_{mid} - \frac{w_s}{2}, \tau_{mid} + \frac{w_s}{2}\right]
\end{align*}
This formula modifies the start and stop times of each event to align with the temporal size of the partitions. The midpoint of the event remains the same, but the duration is adjusted to match $w_s$.

\begin{definition}
\label{def_1}
Let $p_i = \{t_i, t_{i+1}, \ldots, t_{i+w-1}\}$, where $i \in I$, be a partition. The event $e = [\tau_{1}, \tau_{2}]$, where $e \in E$, is considered close to $p_i$ if:
$$|t_i - \tau_1| < w_s.$$
\end{definition}

\begin{definition}
Given $p_i$ and $e = [\tau_{1}, \tau_{2}] \in E$, with $t_{i+w-1} = t_i + w_s$, we define $p_i \cap e$ and $p_i \cup e$ as follows:
\begin{align*}
p_i \cap e = \begin{cases}
\emptyset,&{\text{if}}\ e \text{ is not close to } p_i \text{ (Def. \ref{def_1})}\\
[\tau_1, t_{i+w-1}], & \text{if } t_i \leq \tau_1 \land  t_i > \tau_1 - w_s \\
[t_i, \tau_2], & \text{if } \tau_1 < t_i \land t_i < \tau_2 \
\end{cases}
\end{align*}

To simplify, we define $I_1 = ]\tau_1 - w_s, \tau_1]$ and $I_2 = ]\tau_1, \tau_2[$.

\begin{align*}
p_i \cup e &= \begin{cases}
[t_i, \tau_2], & \text{if } t_i \in I_1 \\
[\tau_1, t_{i+w-1}], & \text{if } t_i \in I_2 \
\end{cases}
\end{align*}
\end{definition}

Based on the definitions, we can express $op(p_i, e)$ as follows:

\begin{align*}
op(p_i, e) &= \begin{cases}
0,& \text{if $e$ is not close to } p_i \\
\dfrac{t_i + w_s - \tau_1}{\tau_2 - t_i}, & \text{if } t_i \in I_1 \\
\dfrac{\tau_2 - t_i}{t_i + w_s - \tau_1}, & \text{if } t_i \in I_2 \
\end{cases}
\end{align*}

\begin{figure}[ht]
\vskip 0.2in
\begin{center}
\centerline{\includegraphics[width=\columnwidth]{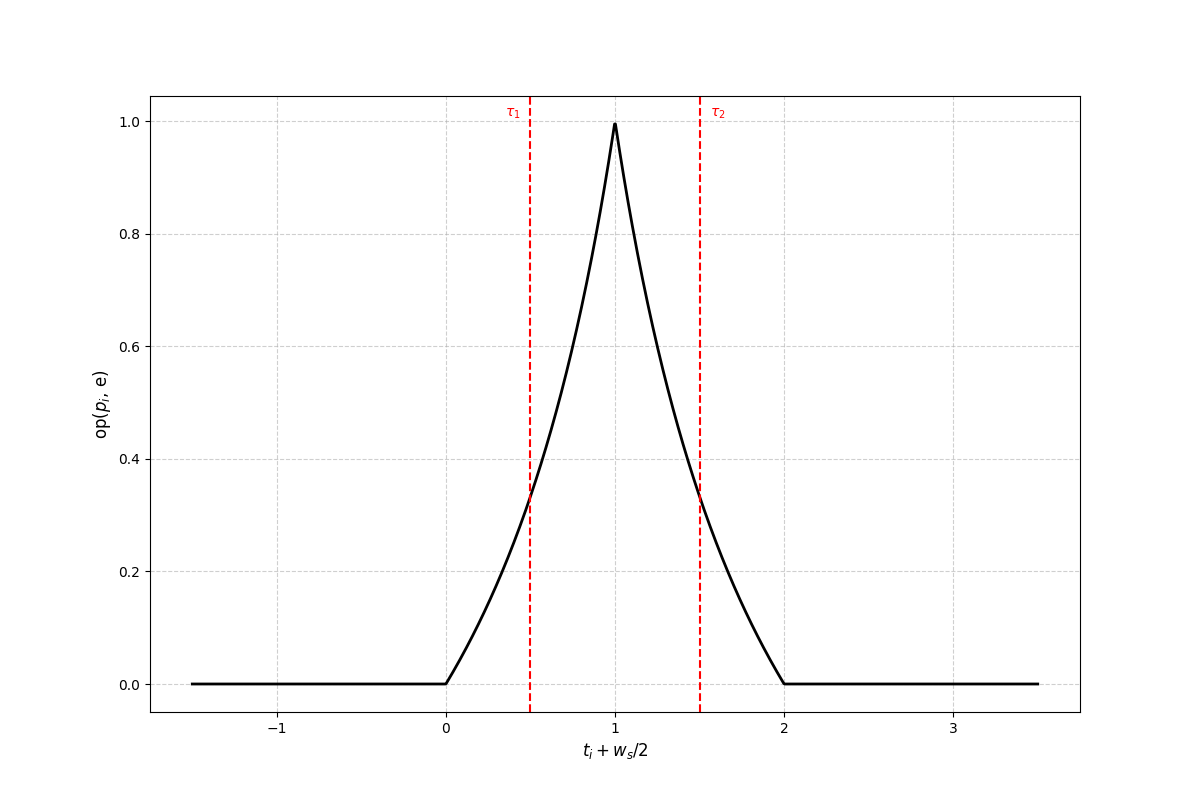}}
\caption{Plot of $op(p_i, e)$ as a function of $t_i + w_s/2$, the middle time of partition $p_i$.}
\label{fig_1}
\end{center}
\vskip -0.2in
\end{figure}

It is readily apparent that when $t_i = \tau_1$, $op(p_i, e)$ equals 1. Similarly, when $t_i = \dfrac{\tau_{1} + \tau_{2}}{2}$, $op(p_i, e)$ equals $\frac{1}{3}$. \cref{fig_1} provides a visual representation of how $op(p_i, e)$ changes with respect to the middle time of $p_i$, denoted as $t_i + \frac{w_s}{2}$. As can be seen from the plot, it is evident that the peak of $op(p_i, e)$ is situated at the midpoint of event $e$.

\begin{definition}
We now define the $op$ value of $p_i$ with respect to all the events $E$ as follows: 
\begin{equation}
\label{eq:1}
op(p_i) = \max_{e\in E} op(p_i, e).
\end{equation}  
\end{definition}

This definition suggests that $op(p_i)$ is equivalent to $op(p_i, e)$, where $e$ denotes the event that is closest to $p_i$. This is because the event closest to $p_i$ exhibits the maximum overlap with the partition $p_i$.

\begin{definition}
We introduce a new function $\pi$ defined as follows:
\begin{align*}
\pi: \mathcal{V} &\rightarrow [0 , 1] \\
v_i &\mapsto \pi(v_i) = op(p_i) 
\end{align*}
where $p_i$ is a partition corresponding to vector $v_i$.
\end{definition}

The function $\pi$ is characterized as a black-box function, which means that there is no explicit mathematical formulation available for it. It's important to note that by selecting a sufficiently large value for $w$, we can ensure a one-to-one correspondence between each $v_i$ in $\mathcal{V}$ and a unique $p_i$ in $\mathcal{P}$. This is without loss of generality, as assuming that the function $o$ is bijective allows us to associate $v_i$ with $p_i$ having the same index in the definition of the $\pi$ function. This assumption of bijectivity simplifies the association process. The function $\pi$ plays a pivotal role in the event detection process, a topic that will be further elaborated in the following sections.

\section{Principle of Event Detection}
\label{principle_of_event_detection}

The principle of event detection, based on overlapping partitions $\mathcal{P}$ and the function $\pi$, centers around identifying a function $f$ with a well-defined mathematical expression that accurately approximates the function $\pi$. A straightforward approach involves training a feed-forward neural network $f$ on the set $\{(v_i, \pi(v_i)) | i\in  I_{train} \subset I\}$.

The approximation error for $f$ is computed as follows:
$$\varepsilon(f)=\frac{1}{|I_{train}|}\sum_{i \in  I_{train}  } \mathcal{L}(f(v_i),\pi(v_i))$$
In this equation, $\mathcal{L}$ denotes the loss function, which measures the disparity between the approximation value $f(v_i)$ and the ground truth value $\pi(v_i)$. In regression problems, the mean squared error (MSE) is commonly employed as the loss function \cite{166bbdd6-5823-3cf8-b004-27f4680753a4}.

Once trained, $f$ can be applied to any vector $v_k$ for $k\in I_{test} \subset I$ to estimate $\pi(v_k)$, equivalent to $op(p_k)$. The peaks in $op(p_k)$ for $k\in I_{test}$ should align with the mid-times of the predicted events, as previously illustrated in \cref{fig_1}. These peaks are identified to extract the predicted events, characterized by the intervals:
$$e_q = \left[\tau_q - \frac{w_s}{2}, \tau_q + \frac{w_s}{2}\right]$$
In this equation, $\tau_q$ denotes the mid-time of the $q$-th peak.

\begin{theorem}
\label{thm:universal}
If $T$ and $T^{-1}$ are continuous, then there exists a feed-forward neural network $f \in \Sigma^{r}(\Psi)$ that utilizes a squashing function $\Psi$. This network can approximate the function $\pi$ from $\mathcal{V}$ to $[0,1]$ with arbitrary precision, given a sufficient number of hidden units $Q \in \mathbb{N}$.

A squashing function $\Psi$ is a type of function that compresses the input into a smaller range. In neural networks, squashing functions, such as the sigmoid and hyperbolic tangent (tanh), serve as activation functions. These functions transform any input value into a range between 0 and 1 (for sigmoid) or between -1 and 1 (for tanh). 

Here, $\Sigma^{r}(\Psi)$ represents a set of single hidden layer feed-forward neural networks defined as follows:
\[ \{ f: \mathbb{R}^{r} \rightarrow \mathbb{R}: f(x) = 
\sum_{j=1}^{Q} \beta_j \Psi(A_j(x)) \} \] 

where $x \in \mathbb{R}^{r}$, $\beta_j \in \mathbb{R}$, and $A_j \in \mathbf{A}^{r}$. The function $A_j(x) = w_j \cdot x + b_j$, with $w_j \in \mathbb{R}^{r}$ and $b_j \in \mathbb{R}$. The parameters $w_j, b_j$, and $\beta_j$ correspond to the network weights.
\end{theorem}

We provide the proof for Theorem \ref{thm:universal} in \cref{proofs}.

Under the continuity of both $T$ and $T^{-1}$, this theorem ensures that the function $\pi$ can be effectively approximated by a feed-forward network $f \in \Sigma^{r}(\Psi)$ with enough hidden units, achieving any desired precision. This establishes the foundation for the reliable and effective method of accurately detecting a wide range of events in time series data, as discussed in the earlier principles of detection.
We note that the theoretical guarantees of this method depend on the continuity of both the time series and its inverse. However, in practical scenarios where continuity may not hold, the method still performs well based on the empirical success of neural networks in various applications. Neural networks have shown their ability to learn and approximate complex patterns, even without strict continuity assumptions. Therefore, while the theory assumes continuity, the adaptability and learning capability of neural networks allow this framework to handle cases where continuity is relaxed.
Our proof demonstrates the value of this method for applications in signal processing, anomaly detection, and prediction across various fields, such as finance, medicine, and engineering.

\subsection{Practical Considerations and Implementations}

In practical applications, feed-forward neural networks for regression tasks often introduce noise into predictions due to factors such as complex data, overfitting, and training limitations. In our scenario, the noise in predictions generated by $f \in \Sigma^{r}(\Psi)$ can pose challenges when estimating peak locations, potentially leading to false events.

To address this issue, one universally effective method is to smooth the approximation f by convolving it with a Gaussian kernel. This convolution operation attenuates high-frequency noise while preserving the underlying shape of f \cite{smoothing}, resulting in more accurate and reliable peak location estimation and a reduction in false events.

The extent of smoothing is governed by the standard deviation $\sigma$. Achieving optimal smoothing outcomes requires selecting the optimal standard deviation. This is accomplished using an optimization algorithm that determines the value of $\sigma$ that maximizes the F1-Score.

Additionally, optimizing the peak threshold is essential to complement the smoothing process. The peak threshold determines which values in the smoothed f are considered as peaks. A suitable threshold ensures a balance between capturing ground truth peaks and minimizing the inclusion of noise-induced peaks. Similar to the standard deviation, optimizing the peak threshold is crucial for accurate peak detection results.

For further details and comprehensive discussions, the reader is referred to the additional materials presented in \cref{noise_reduction}.

\section{Numerical Study}

In this section, we conduct a comprehensive evaluation of the method's effectiveness on two challenging and imbalanced datasets. The first dataset focuses on credit card fraud detection \cite{Benchaji2021-cs}, while the second dataset involves the detection of bow shock crossings in space physical time series. We employ the F1-Score metric, which adeptly balances precision and recall, making it particularly suited for assessing performance on imbalanced datasets where the minority class holds significant interest.

We benchmark the F1-Score results against state-of-the-art metrics obtained by diverse architectures, as reported in reputable literature \cite{Alarfaj2022-id, Varmedja2019-yb, Ileberi2022-qv, Cheng2022-pt}. All results and materials from this section are available on our GitHub repository for reproducibility.

\subsection{Fraud Detection}

\subsubsection{Setup}

The credit card fraud detection dataset labels each one-second time step as either 0 or 1, indicating the absence or presence of fraud, respectively. However, the method requires fraud instances to be represented as a list of time intervals. To reconcile this, we transform the labeled time steps corresponding to fraud occurrences into a list of time intervals. For each time step $c_{1_q}$ where the label is 1 (indicating fraud), with the total number of frauds denoted as $n_b$, we define an interval as follows:
\[ e_q = \left[c_{1_q} - \frac{w_s}{2}, c_{1_q} + \frac{w_s}{2}\right] \]

Subsequently, we construct the list of fraud intervals:
\[ E = \{ e_q \mid q = 1, 2, \dots, n_b \} \]

Given that the fraud dataset has time steps of one second, i.e., $s=1$ second, and considering that the deep learning methods we are comparing with are based on binary classification (predicting the presence or absence of fraud for each second), we need to ensure a fair comparison with our method. Therefore, we consider the temporal size $w_s$ of fraud to be 1 second, which is specified by $w_s = (w-1) \cdot s$. Thus, we must set $w = 2$. For our method, we use an FFN with a single hidden layer containing $Q=20$ neurons. The activation function $\Psi$ is set to sigmoid and the partition size $w$ is chosen to be 2.

\subsubsection{Comparison with and without Data Balancing (SMOTE)}

We compare the method with deep learning approaches that either employ or do not employ any data balancing technique, such as Synthetic Minority Over-sampling Technique (SMOTE) \cite{DBLP:journals/corr/abs-1106-1813}. The benchmark methods include a Convolutional Neural Network (CNN) \cite{Alarfaj2022-id} without SMOTE and Feed-Forward Neural Networks (FFN) \cite{Varmedja2019-yb, Ileberi2022-qv} with SMOTE.

\begin{table*}[t]
\caption{Comparative Analysis of Credit Card Frauds Detection Methods}
\label{tab:results_combined}
\centering
\begin{small}
\begin{sc}
\begin{tabular}{lccccc}
\toprule
Method & Data Balancing & Number of Parameters & Precision & Recall & F1-Score \\
\midrule
CNN \cite{Alarfaj2022-id} & No & 119,457 & 0.89 & 0.68 & 0.77 \\
FFN+SMOTE \cite{Varmedja2019-yb} & Yes & 5,561 & 0.79 & 0.81 & 0.80 \\
FFN+SMOTE \cite{Ileberi2022-qv} & Yes & N/A & 0.82 & 0.79 & 0.81 \\
Ours & No & \textbf{1,201} & 0.98 & 0.74 & \textbf{0.85} \\
\bottomrule
\end{tabular}
\end{sc}
\end{small}
\end{table*}

The method demonstrates superior performance (F1-Score) compared to the aforementioned methods. Furthermore, our approach exhibits a substantial reduction in parameters, with 1,201 parameters as opposed to 119,457 in the CNN-based method and 5,561 in one of the FFN-based methods. This significant parameter reduction not only indicates computational efficiency but also renders our method well-suited for scenarios with limited computational resources. Our method surpasses both competing methods that used SMOTE in terms of precision and F1-Score, achieving high performance even without using SMOTE for data balancing.

\begin{figure*}[t]
\centering
\begin{minipage}{0.45\textwidth}
\includegraphics[width=\linewidth]{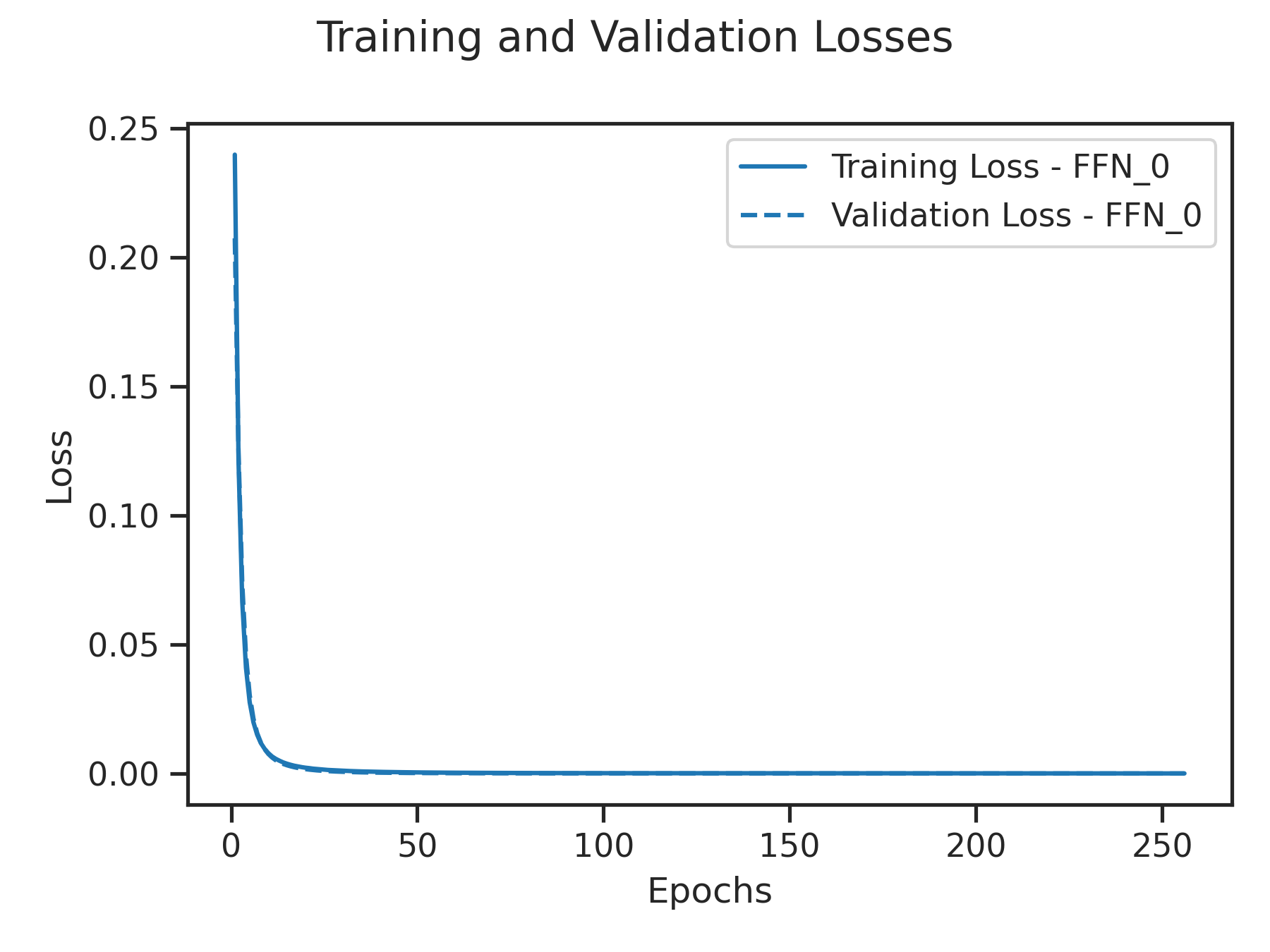}
\caption{The training loss and validation loss of the FFN trained (FFN\_0) on the frauds detection case.}
\label{fig:losses_ccf}
\end{minipage}
\hfill
\begin{minipage}{0.45\textwidth}
\includegraphics[width=\linewidth]{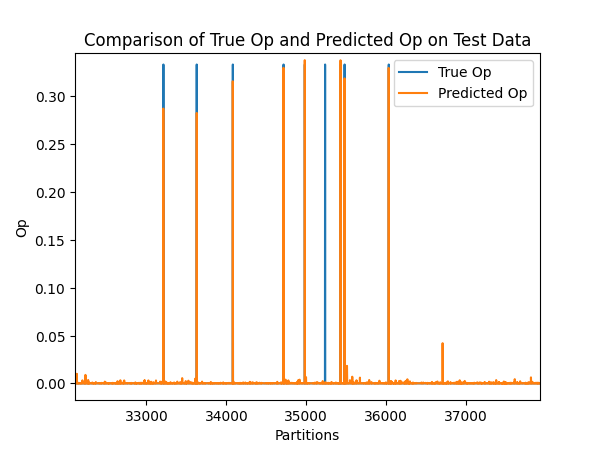}
\caption{Comparison between predicted and ground truth $op$ values on the frauds detection case.}
\label{fig:op_ccf}
\end{minipage}
\end{figure*}

Figure \ref{fig:losses_ccf} showcases the training loss and validation loss of FFN during the training process on fraud detection. The low losses observed in both the training and validation phases indicate that the network has successfully learned the underlying patterns, justifying the obtained good metrics.

Furthermore, Figure \ref{fig:op_ccf} zooms in on the comparison between the predicted $op$ values and the true $op$ values in the credit card fraud case. As expected, the shape of the predicted values aligns well with the true values, showing minimal fluctuations, and sometimes additional peaks appear. This result highlights the effectiveness of using a Gaussian filter to attenuate these fluctuations and a peak detection threshold to eliminate false peaks, thereby leading to improved accuracy of the predictions and justifying the excellent metrics obtained. For more discussions on this, see Section \ref{add_results}.

\subsection{Bow Shock Crossings Detection}

In this section, we extend the evaluation of the method to a distinct imbalanced dataset explicitly designed for plasma's physical time series with bow shock crossing events. The dataset is derived from Mars Express spacecraft (MEX) data using the AMDA software \cite{Genot2021-sv}, with ground truth annotations for bow shock crossings provided by \cite{Hall2016-kg}.

We compare the method with a study by \cite{Cheng2022-pt}, where the authors developed a deep learning approach for bow shock crossing detection using Cassini data and ResNet18, a well-known deep CNN architecture. Despite the entirely different nature of our dataset, focusing on time series data rather than images, we choose to compare with \cite{Cheng2022-pt} due to the absence of other bow shock detection studies using deep learning on time series data. We aim to demonstrate that our method, with a minimal number of parameters, can achieve metrics comparable to state-of-the-art architectures like ResNet18.

\subsubsection{Setup}

In configuring our approach, we intentionally selected a partition size $w$ of 76. This decision aligns with the empirical understanding that a bow shock event typically spans 5 minutes, as indicated in \cite{Hall2016-kg}, where criteria are assessed within a 5-minute timeframe to validate bow shock crossings. With our dataset's sampling rate denoted by $s$ and set at 4 seconds, the event duration $w_s$ is precisely calculated as $(76-1)\cdot 4 = 300$ seconds, corresponding to the acknowledged 5-minute duration.

In this specific configuration, we employ a feed-forward neural network featuring a single hidden layer. The hidden layer comprises $Q=20$ neurons and utilizes a sigmoid function as a squashing function.

\subsubsection{Performance Evaluation and Comparative Analysis}

\begin{table*}[t]
\caption{Comparative Analysis of Bow Shock Crossings Detection Methods}
\label{tab:results_bs}
\centering
\begin{small}
\begin{sc}
\begin{tabular}{lccccc}
\toprule
Method & Number of Parameters & Precision & Recall & F1-Score \\
\midrule
ResNet18 \cite{Cheng2022-pt} & 29,886,979 & 0.99 & [0.83 , 0.88] & [0.91 , 0.94] \\
Ours & \textbf{6,121} & 0.95 & 0.96 & \textbf{0.95} \\
\bottomrule
\end{tabular}
\end{sc}
\end{small}
\end{table*}

The method outperforms the ResNet18-based method. Moreover, the method's efficiency is evident in its significantly lower number of parameters, demonstrating its ability to achieve high performance with low complexity.

\begin{figure*}[t]
\centering
\begin{minipage}{0.45\textwidth}
\includegraphics[width=\linewidth]{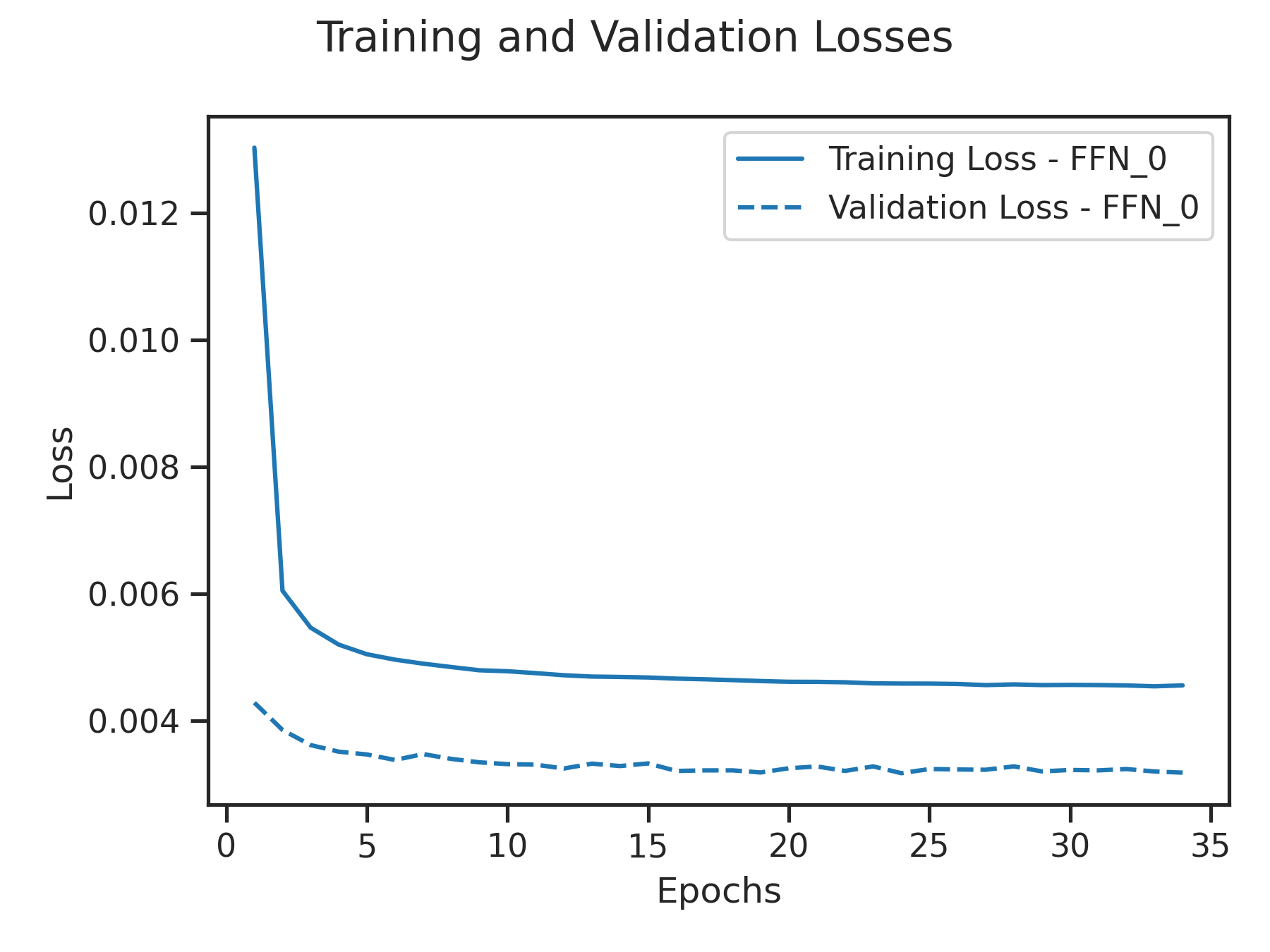}
\caption{The training loss and validation loss of the FFN trained (FFN\_0) on the bow shock crossings case.}
\label{fig:losses_bs}
\end{minipage}
\hfill
\begin{minipage}{0.45\textwidth}
\includegraphics[width=\linewidth]{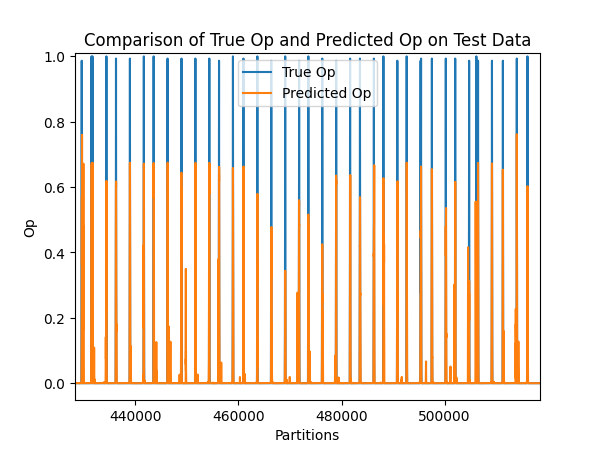}
\caption{Comparison between predicted and ground truth $op$ values on the bow shock crossings case.}
\label{fig:op_bs}
\end{minipage}
\end{figure*}

Figure \ref{fig:losses_bs} showcases the training loss and validation loss of FFN during the training process of the Bow Shock case.

Figure \ref{fig:op_bs} zooms in on the comparison between the predicted $op$ values and the true $op$ values on the bow shock case. In this case, we see a better matching between predicted values with the true values than in the previous case, which justifies a better F1-Score.

Here is the corrected and improved version of the "Conclusion" section:

\section{Conclusion}

In this paper, we have presented a novel deep learning supervised method for event detection in multivariate time series data, leveraging a regression-based approach instead of traditional classification. We have established a rigorous theoretical foundation, making this method a versatile framework capable of detecting a wide range of events, including change points, frauds, anomalies, and more. By demonstrating its universality under mild assumptions about the continuity of multivariate time series, we have established its ability to identify events with arbitrary precision.

Our framework not only excels in theoretical considerations but also exhibits practical efficacy. With a minimal number of trainable parameters, our approach outperforms existing deep learning methods on real-world imbalanced datasets, particularly in fraud detection and bow shock crossing identification. These practical validations underscore the effectiveness and relevance of our framework across various domains, establishing it as a compelling solution for event detection in diverse fields.

However, it is essential to acknowledge certain limitations in our framework, particularly in the context of multi-class event detection, where complexities may arise. To address these limitations, we plan to extend the evaluation of our framework to diverse datasets from various fields, aiming to achieve state-of-the-art metrics compared to other methods. This broader testing approach will allow us to further demonstrate the advantages of our method, including its minimal parameter requirements, which not only reduce computational resources but also contribute to a more sustainable and environmentally friendly solution. Additionally, our ongoing efforts involve enhancing the framework's capabilities to predict events of varying durations, addressing a notable limitation in our current approach, which focuses solely on predicting the midpoint of events with fixed durations. This advancement represents a significant improvement in the versatility and practical applicability of our framework.

Overall, the proposed method represents a significant step forward in event detection for multivariate time series data. Its theoretical rigor, practical efficacy, and minimal parameter requirements make it a compelling solution for a wide range of applications. We believe that our method has the potential to revolutionize event detection in various fields and will continue to explore its capabilities and expand its applicability in future work.

\bibliography{example_paper.bib}
\bibliographystyle{icml2024}

\newpage
\appendix
\onecolumn
\section{Proofs}
\label{proofs}

\begin{proof}
\label{proof:1}
To prove Theorem \ref{thm:universal}, we first need to establish the continuity of the following three functions: $o$, $op$, and $\pi$. But before delving into that, let's provide some definitions.

\begin{definition}
We define a distance $d_{\mathcal{P}}$ as follows:
\begin{align*}
    d_{\mathcal{P}}: \mathcal{P} \times \mathcal{P} &\rightarrow \mathbb{R} \\
    (p_i, p_j) &\mapsto d_{\mathcal{P}}(p_i, p_j) = \lvert t_i - t_j \rvert
\end{align*}
\end{definition}

\begin{definition}
We define a distance $d_{\mathcal{V}}$ as follows:
\begin{align*}
    d_{\mathcal{V}}: \mathcal{V} \times \mathcal{V} &\rightarrow \mathbb{R} \\
    (v_i, v_j) &\mapsto d_{\mathcal{V}}(v_i, v_j) = \sqrt{\sum_{m=1}^{w} \sum_{k=1}^f (T(t_{j+m-1})[k] - T(t_{i+m-1})[k])^2}
\end{align*}
\end{definition}

\begin{definition}
We define a distance  $d_{T}$ as follows:
\begin{align*}
    d_{T}: \mathbb{R} \times \mathbb{R} &\rightarrow \mathbb{R} \\
    (t_i, t_j) &\mapsto d_{T}(t_i, t_j) = \sqrt{\sum_{k=1}^f (T(t_{j})[k] - T(t_{i})[k])^2}
\end{align*}
\end{definition}

Let $i, j \in I$ and $m \in \mathbb{N}$, such that $0 < m \leq w$ and $0 < i < j$.

\textbf{Proving the Continuity of $o$:}

We have $\lvert t_{j+m-1} - t_{i+m-1} \rvert = \lvert t_j - t_i \rvert = (j - i) \cdot s$. Given $\epsilon = s > 0$, without loss of generality, assume $s \ll 1$ (assuming $N$ is sufficiently large). Then, there exists $\delta = (j - i)\epsilon + \epsilon$ such that $\lvert t_{j+m-1} - t_{i+m-1} \rvert < \delta$. If we assume that the time series $T$ is continuous, then $\lvert t_{j+m-1} - t_{i+m-1} \rvert < \delta$ implies $d_{T}(t_{j+m-1}, t_{i+m-1}) < \epsilon$.

For $p_i, p_j \in \mathcal{P}$, let $v_i = o(p_i)$ and $v_j = o(p_j)$. From the definition:
\begin{align*}
    d_{\mathcal{V}}(v_i, v_j) &= \sqrt{\sum_{m=1}^{w} \sum_{k=1}^f (T(t_{j+m-1})[k] - T(t_{i+m-1})[k])^2}.
\end{align*}
If $d_{T}(t_{j+m-1}, t_{i+m-1}) < \epsilon$, then we can bound $d_{\mathcal{V}}(v_i, v_j)$ as follows:
\begin{align*}
    d_{\mathcal{V}}(v_i, v_j) < \sqrt{w} \epsilon.
\end{align*}

Let $\epsilon_1 = \sqrt{w} \epsilon$. Choose $\delta = (j - i) \frac{\epsilon_1}{\sqrt{w}} + \frac{\epsilon_1}{\sqrt{w}}$. Now, if $d_{\mathcal{P}}(p_i, p_j) < \delta$, it implies $d_{\mathcal{V}}(v_i, v_j) < \epsilon_1$. Thus, we can conclude that $o$ is continuous.

\textbf{Proving the Continuity of $op$:}

For $p_i, p_j \in \mathcal{P}$, we have $d_{\mathcal{P}}(p_i, p_j) = (j-i) \cdot s$. Assuming that $s$ is very small (given $N$ is sufficiently large), without loss of generality, let's consider $s \ll 1$. Under this assumption:

\begin{itemize}
    \item $p_i$ and $p_j$ share the same closest event $e$. Consequently, we express $op(p_i)$ as $op(p_i, e)$ and $op(p_j)$ as $op(p_j, e)$.
    \item If $t_i \in I_1$, then $t_j \in I_1$, and if $t_i \in I_2$, then $t_j \in I_2$.
\end{itemize}

For $t_j, t_i \in I_1$, the bound for $op(p_j) - op(p_i)$ is given by:
\[op(p_j) - op(p_i) < \frac{2d_{\mathcal{P}}(p_j, p_i)}{w_s}\]

Similarly, for $t_j, t_i \in I_2$, the bound is:
\[op(p_j) - op(p_i) > -\frac{2d_{\mathcal{P}}(p_j, p_i)}{w_s}\]

In summary:
\[|op(p_j) - op(p_i)| < \frac{2d_{\mathcal{P}}(p_j, p_i)}{w_s}\]

This relation is the Lipschitzian relation, implying the continuity of $op$ \cite{Cobzas2019-xt}.

\textbf{Proving the Continuity of $\pi$:}

Let $(v_{i_n})n$ be a sequence that converges to $v_i$, where $v_i, v{i_n} \in \mathcal{V}$. This can be formally written as:
$$\lim_{{n \to \infty}} v_{i_n} = v_i$$
This means that as $n$ approaches infinity, the sequence $(v_{i_n})n$ converges to $v_i$. In other words, the values of $v{i_n}$ get arbitrarily close to $v_i$ as $n$ gets larger and larger. These assignments are valid since the function $o$ is continuous. For more on convergence sequences with continuous functions, please refer to \cite{Hoffman1975-tw}.

Let's define $v_i, v_{i_n}$ as follows:
\[
v_i = \begin{pmatrix}
    x_i \\
    \vdots \\
    x_{i+w-1}\\
\end{pmatrix},
v_{i_n} = \begin{pmatrix}
    x_{i_n} \\
    \vdots \\
    x_{(i+w-1)_n}\\
\end{pmatrix}
\]
Where $x_{(i+k)_n}, x_{i+k} \in \mathbb{R}^f, 0 \leq k \leq w-1$. We have $\lim_{{n \to \infty}} v_{i_n} = v_i \implies \lim_{{n \to \infty}} x_{(i+k)_n} = x_{i+k}, 0 \leq k \leq w-1$ \cite{Rudin1967}.
Since $o$ is bijective, we can associate, for $v_i, v_{i_n}$ respectively, the partitions $p_i, p_{i_n}$ that can be defined as follows:
$$p_i = \{T^{-1}(x_{i+k}), 0 \leq k \leq w-1\}$$
$$p_{i_n} = \{T^{-1}(x_{(i+k)_n}), 0 \leq k \leq w-1\}$$
We have for $0 \leq k \leq w-1$ that $\lim_{{n \to \infty}} x_{(i+k)_n} = x_{i+k}$ then if $T^{-1}$ is continuous, we can deduce that:
$$\lim_{{n \to \infty}} p_{i_n} = p_i$$ 
Since $op$ is continuous, then:
$$\lim_{{n \to \infty}} p_{i_n} = p_i \implies \lim_{{n \to \infty}} op(p_{i_n}) = op(p_i)$$
Finally, from the definition of the function $\pi$, we can write that $\lim_{{n \to \infty}} v_{i_n} = v_i \implies \lim_{{n \to \infty}} \pi(v_{i_n}) = \pi(v_i)$, thus $\pi$ is continuous \cite{Rudin1967}.

We have $\mathcal{P}$ as a finite set, implying it is compact. Since $o$ is continuous, the image of $\mathcal{P}$ under $o$,  $\mathcal{V}$, is also compact \cite{Rudin1967}. Therefore, $\pi$ is continuous over the compact set $\mathcal{V}$. According to Theorem 2.4 in \cite{Hornik1989-dg}, we can assert that a feed-forward neural network $f \in \Sigma^{r}(\Psi)$, utilizing a squashing function $\Psi$, can accurately approximate the function $\pi$ with any desired degree of precision $\lambda$. In their proof, the authors choose the number of hidden units $Q$ such that $\frac{1}{Q} < \dfrac{\lambda}{2}$, indicating that a larger value of $Q$ is required to achieve an excellent approximation.
\end{proof}

\section{Practical Considerations and Implementations}
\label{noise_reduction}

\subsection*{Post-Processing for Noise Reduction}

In practical scenarios, the noise in predictions generated by $f \in \Sigma^{r}(\Psi)$ can pose challenges when estimating peak locations, potentially leading to false events. For this reason, it is often necessary to smooth the predicted values $f$ by convolving them with a Gaussian kernel $G_{\sigma}$, characterized by a standard deviation $\sigma$.

The predicted values generated by $f$ are defined as follows:
\[
P_f = \{k \in I_{\text{test}}, f(v_k)\}
\]
By definition (cf. \cref{principle_of_event_detection}), the vector $v_k$ is associated with partition $p_k$, and $op(p_k) = f(v_k)$.

The convolution operation is defined as follows:
\[
P_G[k]=\dfrac{\sum_{x=-r}^{r} P[k-x] \cdot G_{\sigma}[x]}{\sum_{x=-r}^{r} G_{\sigma}[x]}
\]
Where $r \in \mathbb{N}$ is a radius of the Gaussian filter. The Gaussian kernel $G_{\sigma}$ is defined as follows:
\[
G_{\sigma}[x] = \frac{1}{\sqrt{2\pi}\sigma} e^{-\frac{x^2}{2\sigma^2}}
\]
The normalization factor $\dfrac{1}{\sum_{k=-r}^{r} G_{\sigma, r}[k]}$ is applied to ensure that the sum of the normalized kernel values equals 1. This normalization step preserves the overall amplitude of the predicted values during the convolution operation.

The Gaussian function is known for its property of being infinitely differentiable, which means that its derivatives of all orders exist and are continuous. Consequently, when a Gaussian kernel is convolved with a function, the resulting function also inherits this property of infinite differentiability \cite{richard:hal-00519555}. This characteristic simplifies the task of identifying local maxima or peaks.

To distinguish the peaks that most likely represent true events, it is common to introduce a threshold on the peak height values, denoted as $h$. Peaks with values above the threshold are identified as predicted events. Similar to the standard deviation, optimizing the peak threshold is crucial for accurate peak detection results.

\subsection*{Computing Predicted Events}

\begin{figure}[h!]
    \vskip 0.2in
    \begin{center}
        \centerline{\includegraphics[width=\textwidth]{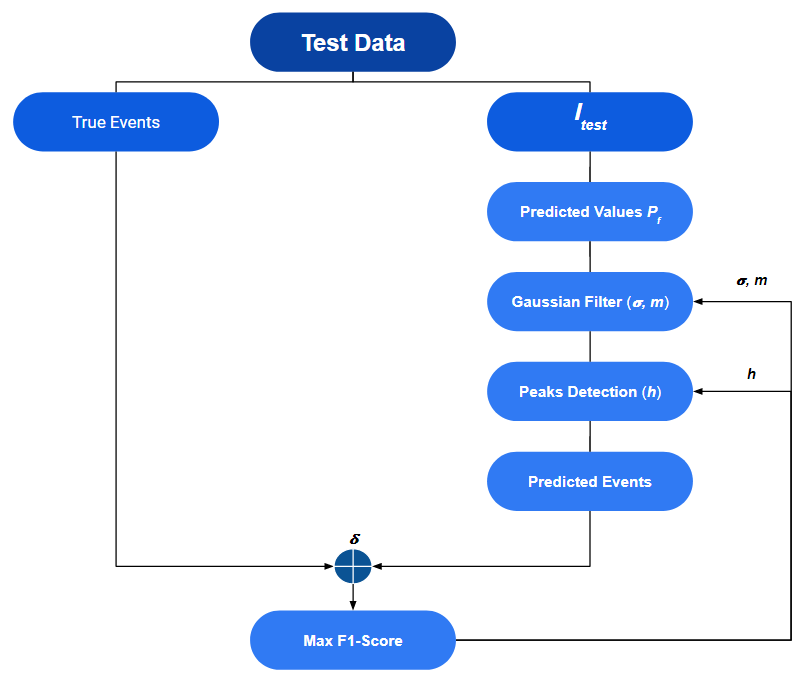}}
        \caption{Overview of the process for computing predicted events, involving smoothing, peak identification, comparison with actual events, and performance evaluation. The figure illustrates the iterative optimization of parameters for enhanced event detection.}
        \label{fig_2}
    \end{center}
    \vskip -0.2in
\end{figure}

To compute the predicted events, we employ the following process that involves the following steps (\cref{fig_2}):

\begin{enumerate}
    \item \textbf{Smoothing:} The predicted values undergo a Gaussian filter to reduce noise and eliminate fluctuations. This smoothing process enhances the accuracy of event extraction by reducing false positives caused by noise.
    \item \textbf{Peak Identification:} After smoothing, we identify peaks in the filtered predictions. These peaks correspond to the mid-times of the predicted events, indicating the locations where events are likely to occur.
    \item \textbf{Comparison with Actual Events:} The identified peaks are compared with the actual events in the test set. A predicted event is considered a match if it occurs within a maximum time tolerance $\delta$ of its corresponding actual event. This time tolerance allows for some flexibility in matching the predicted and actual event times, accommodating the inherent temporal uncertainty in labeling the reference events. The maximum time tolerance parameter $\delta$ is a user-defined value. By default, we set $\delta$ to be equal to $w_s$.
    \item \textbf{Performance Evaluation:} The performance is evaluated using the F1-Score metric. Maximizing the F1-Score is the desired outcome, as it requires simultaneously optimizing precision and recall. To maximize the F1-Score, several parameters are fine-tuned. These include the radius $r$ and the standard deviation ($\sigma$) of the Gaussian filter, and the threshold ($h$) used for peak identification. By optimizing these parameters, we can accurately identify the predicted events, leading to improved overall performance of the method. 
\end{enumerate}

\section{Additional Results}
\label{add_results}
\cref{fig:op_ccf} and \cref{fig:op_bs} present a visual comparison between the predicted values and the ground truth values of $op$. The shape of the predicted values aligns well with the ground truth values, and occasional additional peaks may appear. The effectiveness of using a peak threshold to eliminate false peaks, as explained above, is evident in the improved accuracy of predictions.

\begin{figure}[t!]
    \begin{minipage}{0.45\textwidth}
        \centering
        \includegraphics[width=0.9\textwidth]{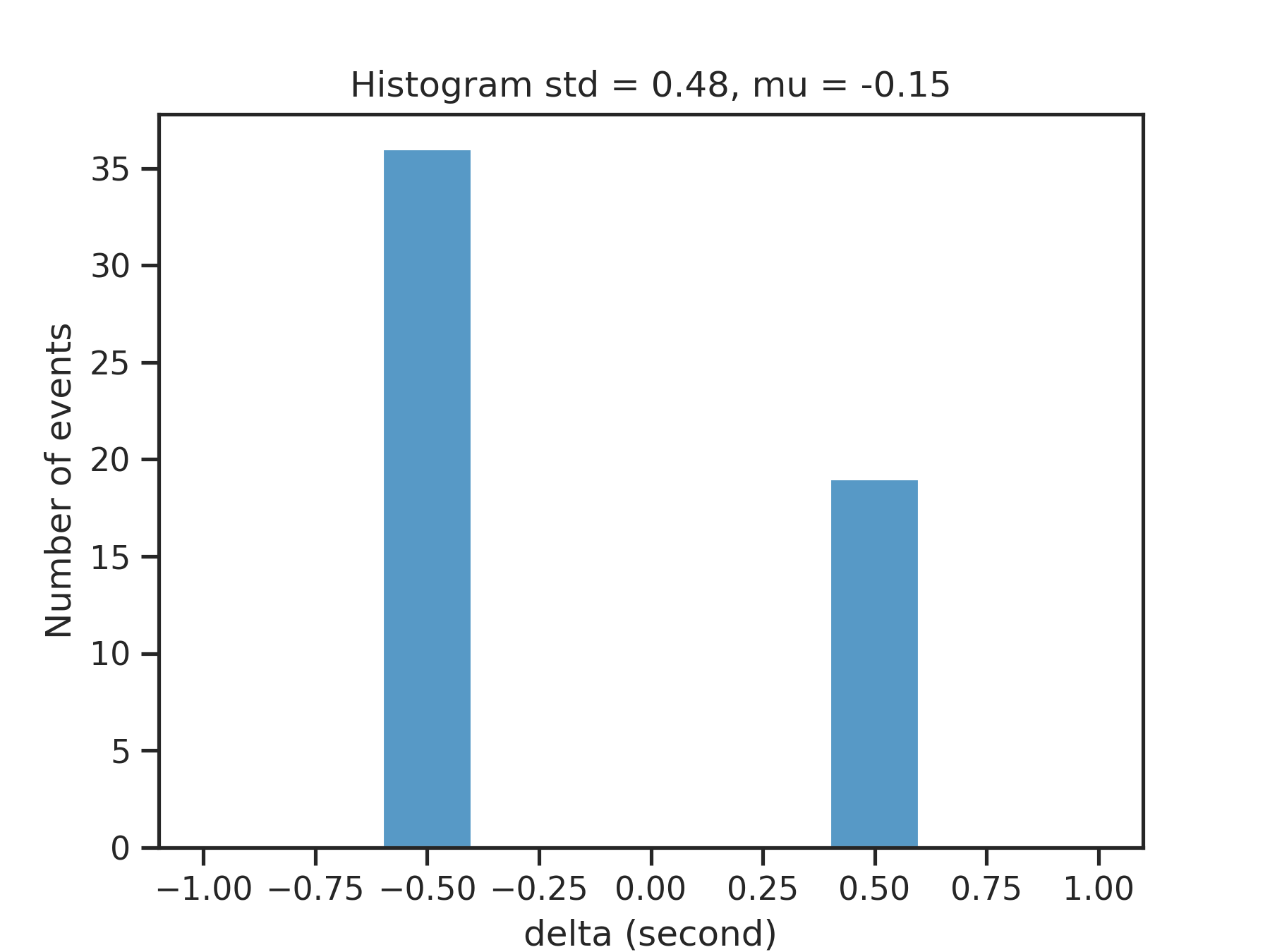}
        \caption{Time difference $\delta(t)$ between predicted and ground truth events on the fraud case.}
        \label{fig:delta_t_ccf}
    \end{minipage}\hfill
    \begin{minipage}{0.45\textwidth}
        \centering
        \includegraphics[width=0.9\textwidth]{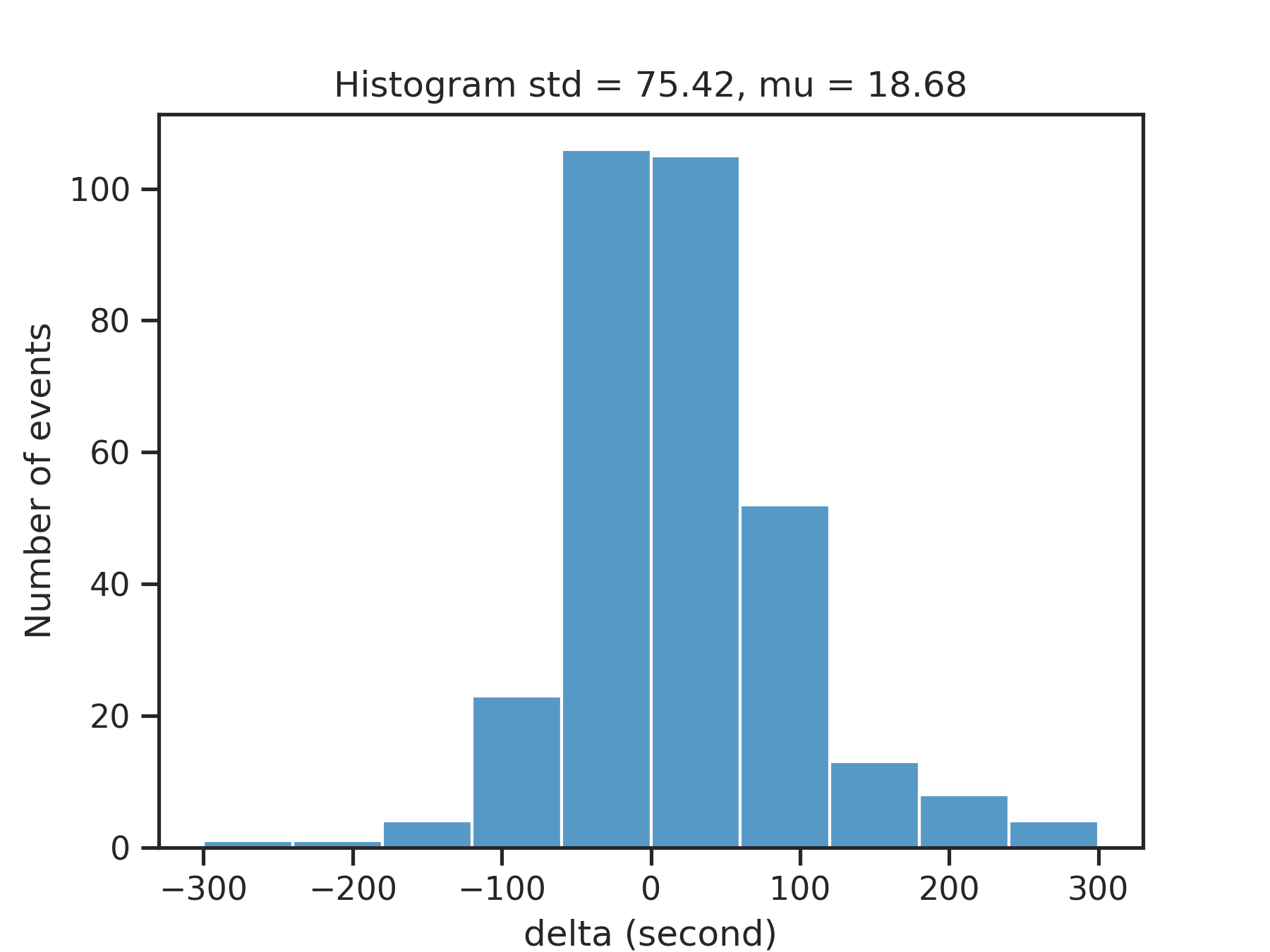}
        \caption{Time difference $\delta(t)$ between predicted and ground truth events on the bow shock case.}
        \label{fig:delta_t_bs}
    \end{minipage}
\end{figure}

Furthermore, \cref{fig:delta_t_ccf} and \cref{fig:delta_t_bs} illustrate the distribution of time differences $\delta(t)$ between predicted events and ground truth events respectively on fraud case and bow shock case. This visualization offers valuable insights into the temporal deviations between predicted and actual event occurrences, facilitating an analysis of the accuracy and precision of our event predictions concerning their temporal alignment. In the fraud case, the mean is equal to -0.15 seconds and a standard deviation of about 0.48 seconds implying that our framework achieves precise event detection. In the bow shock case, the distribution of time differences demonstrates a standard deviation of 75 seconds and a mean of 18 seconds. This indicates that predicted events typically deviate from ground truth events by an average of 18 seconds, with some variations of up to 75 seconds. This level of accuracy is perfectly acceptable for physical applications, especially taking into account the fact that typical planetary plasma instruments have a temporal resolution in the 1-10sec range.

\end{document}